%% file: egpaper_final.tex
\documentclass[10pt,twocolumn,letterpaper]{article}

\usepackage{iccv}
\usepackage{times}
\usepackage{epsfig}
\usepackage{graphicx}
\usepackage{amsmath}
\usepackage{amssymb}
\usepackage{authblk}

\usepackage{graphicx}
\usepackage{amsmath}
\usepackage{amssymb}
\usepackage{booktabs}
\usepackage{multirow}
\usepackage{cite}
\usepackage{rotating}
\usepackage[dvipsnames]{xcolor}

\usepackage{mathrsfs} 
\usepackage{subfigure}
\usepackage{makecell}

\usepackage{amsmath,bm}
\usepackage{float}

\usepackage[breaklinks=true,bookmarks=false]{hyperref}

\iccvfinalcopy 

\def\iccvPaperID{****}

\ificcvfinal\fi

\begin{document}

\title{ELFNet: Evidential Local-global Fusion for Stereo Matching}

\author[1,2]{Jieming Lou}
\author[1]{Weide Liu}
\author[1]{Zhuo Chen}
\author[1]{Fayao Liu}
\author[1]{Jun Cheng\thanks{cheng\_jun@i2r.a-star.edu.sg}}

\affil[1]{Institute for Infocomm Research, A*STAR}
\affil[2]{National Univeristy of Singapore}

\maketitle
\ificcvfinal\thispagestyle{empty}\fi
\input{0_Abstract.tex}
\input{1_Introduction.tex}
\input{2_Related_Work.tex}
\input{3_Methods.tex}
\input{4_Experiments.tex}
\input{5_Conclusion.tex}

\section*{Acknowledgement} 
This work was supported by the Agency for Science, Technology and Research (A*STAR) under its MTC Programmatic
 Funds (Grant No. M23L7b0021).
 
{\small
\bibliographystyle{ieee_fullname}
\bibliography{egbib}
}

\end{document}

%% file: 0_Abstract.tex
\begin{abstract}
Although existing stereo matching models have achieved continuous improvement, they often face issues related to trustworthiness due to the absence of uncertainty estimation. Additionally, effectively leveraging multi-scale and multi-view knowledge of stereo pairs remains unexplored. In this paper, we introduce the \textbf{E}vidential \textbf{L}ocal-global \textbf{F}usion (ELF) framework for stereo matching, which endows both uncertainty estimation and confidence-aware fusion with trustworthy heads.
Instead of predicting the disparity map alone, our model estimates an evidential-based disparity considering both aleatoric and epistemic uncertainties. With the normal inverse-Gamma distribution as a bridge, the proposed framework realizes intra evidential fusion of multi-level predictions and inter evidential fusion between cost-volume-based and transformer-based stereo matching.
Extensive experimental results show that the proposed framework exploits multi-view information effectively and achieves state-of-the-art overall performance both on accuracy and cross-domain generalization. 
 The codes are available at \url{https://github.com/jimmy19991222/ELFNet}.
\end{abstract}

%% file: 1_Introduction.tex
\section{Introduction}
Stereo matching, which aims at estimating the dense disparity map given a pair of rectified images, is one of the most fundamental problems in various applications, such as 3D reconstruction, autonomous driving, and robotics navigation~\cite{hamid2022stereo}. Benefiting from the rapid development of convolutional neural networks, many stereo matching models have achieved promising performance by constructing cost volume and using 3D convolutions~\cite{gwc2019,Wu2019,Xu20,acv2022}. Recently, with the support of transformer, approaches have been proposed to utilize global information using self- and cross-attention mechanisms, bringing an alternative way for the stereo matching~\cite{li2021revisiting, guo2022context}. 

Despite the improved performance, quantifying uncertainty of the stereo matching results has been overlooked.  The frequently occurring overconfident predictions in the existing stereo matching limit the deployment of the algorithms, especially in safety-critical applications. The deep learning models are prone to be unreliable due to the lack of interpretability, especially when facing out-of-domain, low-quality or perturbed samples. Things are even worse in the field of stereo matching where the model is first pretrained in a large scaled synthetic dataset~\cite{sceneflow} 
 and fine-tuned in a much smaller dataset from real-world scenes. This makes uncertainty estimation an essential part of preventing potentially disastrous decisions based on stereo matching results.

\begin{figure}[t]  
\centering
\subfigure[Left image]{
\centering
\includegraphics[width=1.5in]{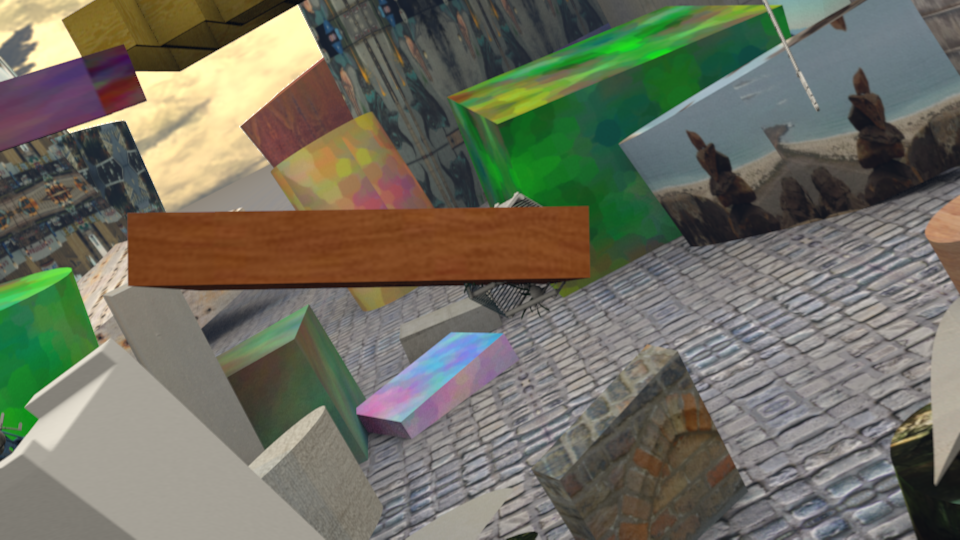}
\label{left image}
}
\subfigure[ELFNet (Ours)]{
\includegraphics[width=1.5in]{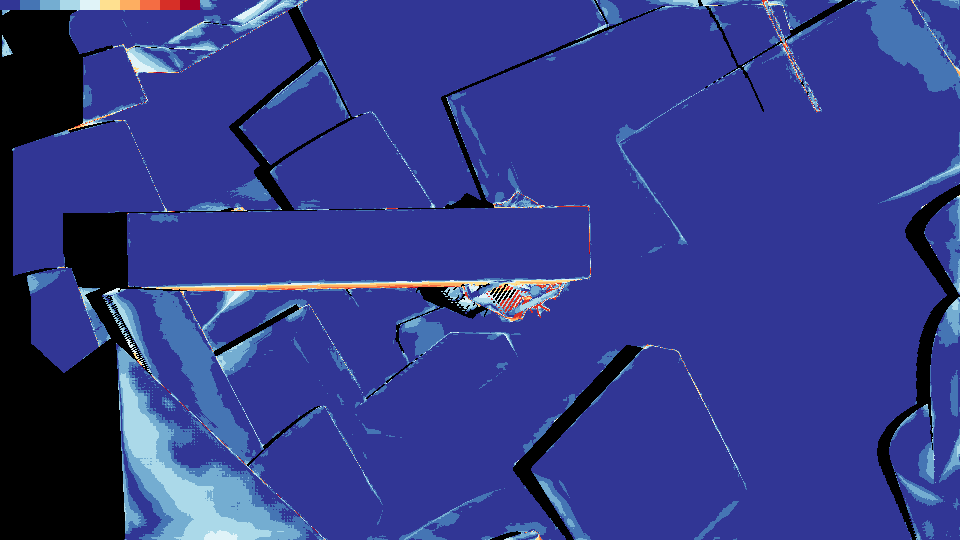}
\label{err_map_elfnet}
}
\subfigure[PCWNet (Cost-volume based)]{
\includegraphics[width=1.5in]{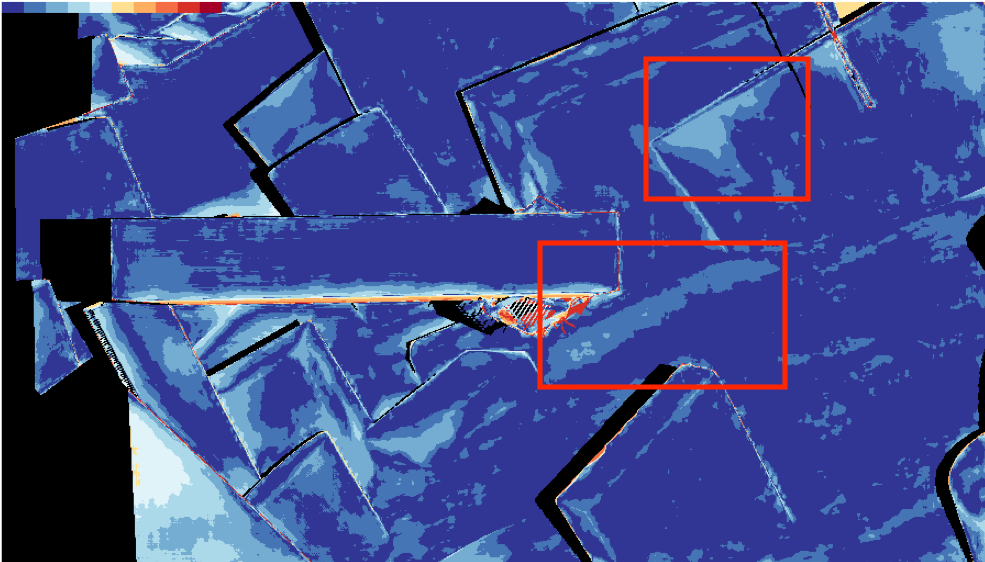}
\label{err_map_pcwnet}
}
\centering
\subfigure[STTR (Transformer based)]{
\centering
\includegraphics[width=1.5in]{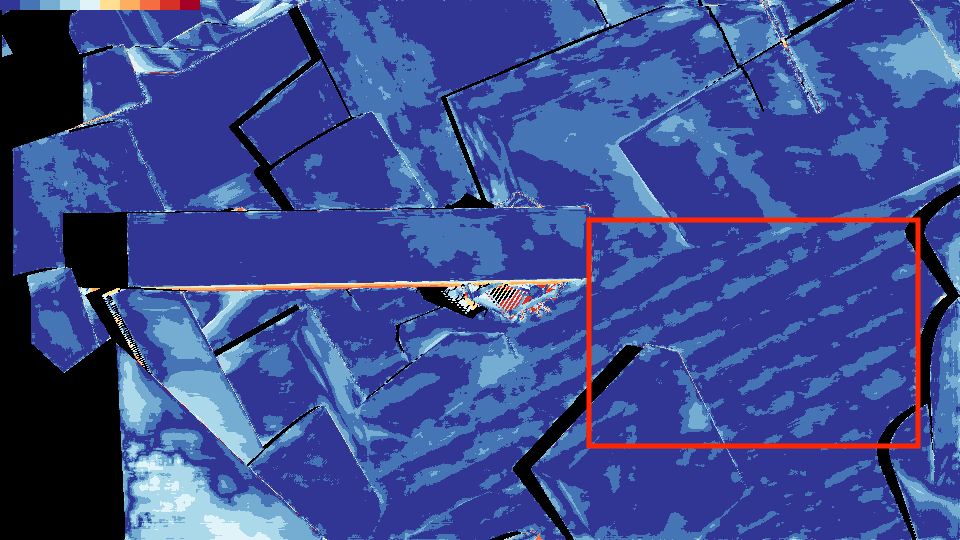}
\label{err_map_sttr}
}
\caption{Error map visualization and comparison of the proposed ELFNet, PCWNet (a cost-volume-based model) and STTR (a transformer-based model) on a case of the Scene Flow~\cite{sceneflow}. The regions colored in blue indicate low error, while the regions in white and red indicate relatively higher error. By fusing cost-volume-based and transformer-based models reliably, our ELFNet significantly reduces the error.}
\label{figure: error map}
\end{figure}

In the meantime, multi-view complementary information widely exists in stereo matching, but it remains a challenge to harness them to improve accuracy effectively and efficiently. 
For instance, multi-scale pyramidal cost volumes are used to offer the coarse-to-fine knowledge obtained from the feature extractor~\cite{Chang18,Wu2019,gwc2019,cheng2020deep,yang2020cost,Shen_2021_CVPR,shen2022pcw}, but the current fusion method fails to consider the uncertainties in different scales, which results in an untrustworthy and incomplete integration. 
In addition, cost-volume-based \cite{acv2022, shen2022pcw} and transformer-based approaches \cite{li2021revisiting} provide entirely different strategies for dealing with stereo pairs: the former aggregates local features with convolutions, and the latter captures global information with transformer for dense matching. We find that these two types of methods complementary to each other.   
 For instance, as shown in the red blocks of Figure~\ref{err_map_pcwnet} and ~\ref{err_map_sttr}, cost-volume-based model is not robust in regions with large illumination changes while transformer-based model does not make full use of complex local textures. In such a scenario, uncertainty estimation is a potential module for endowing a trustworthy fusion strategy among multi-view information to alleviate the error risks without bringing much additional computational load. 

With these motivations in mind, we propose an $\textit{Evidential Local-global Fusion}$ (ELF) framework for stereo matching to kill two birds with one stone (see Figure~\ref{fig:architecture}). The framework enables both uncertainty estimation and reliable fusion by taking advantage of deep evidential learning~\cite{sensoy2018evidential,zhao2019quantifying,amini2020deep}. Specifically, we employ trustworthy heads in each branch of the model to compute the aleatoric and epistemic uncertainties~\cite{der2009aleatory, kendall2017uncertainties} along with the disparity. 
To integrate the multi-scale cost volume information and the complementary information between convolution based and transformer based approaches simultaneously, we propose an intra evidential fusion module and an inter evidential fusion module with a mixture of normal-inverse Gamma distributions (MoNIG)~\cite{de2004wavelet,ma2021trustworthy}. As shown in Figure~\ref{err_map_elfnet}, the proposed ELFNet attains a low disparity error  in most regions by leveraging respective strengths of the cost-volume-based model \cite{shen2022pcw} and the transformer-based model \cite{li2021revisiting} according to the evidence dynamically. 

Our contributions can be summarized as follows:
\begin{enumerate}
\item We introduce deep evidential learning to both cost-volume-based and transformer-based stereo matching to estimate both aleatoric and epistemic uncertainties; \item We propose a novel evidential local-global fusion (ELF) framework, which enables both uncertainty estimation and two-stage information fusion based on evidence; \item We conduct comprehensive experiments, which demonstrate that the designed ELFNet consistently boost the performance in terms of accuracy and cross-domain generalization. \end{enumerate}

%% file: 2_Related_Work.tex
\begin{figure*}[ht]
\centering
\includegraphics[scale=0.25]{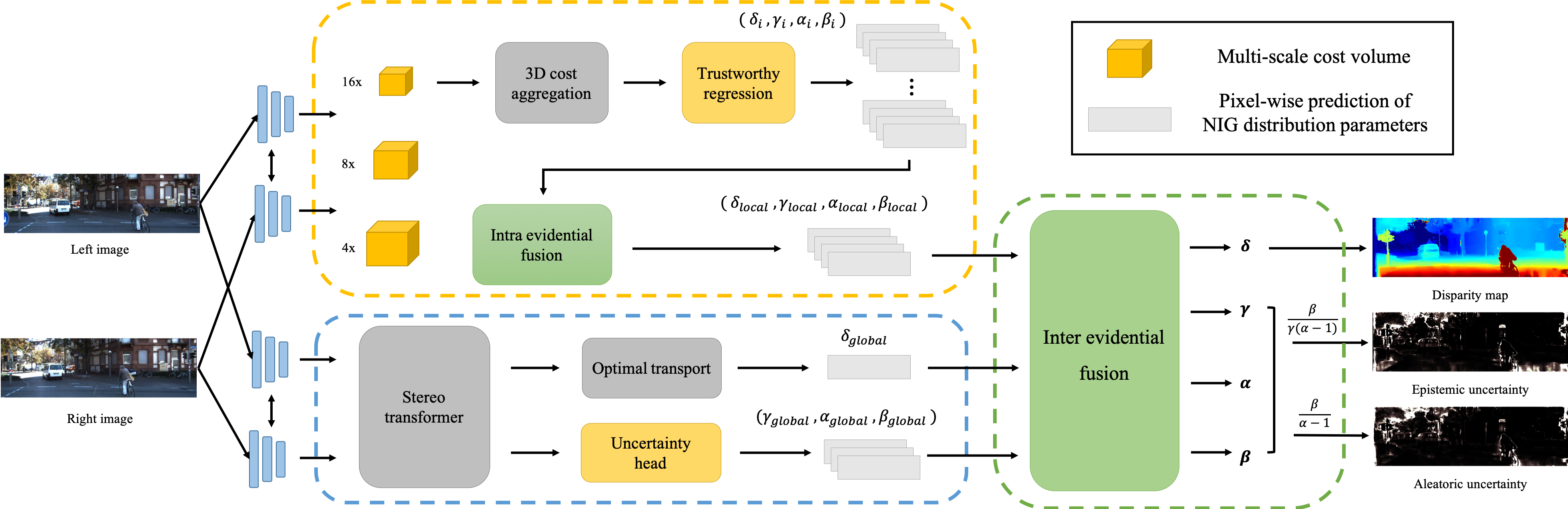}
\caption{Illustration of the proposed evidential local-global fusion (ELF) for stereo matching. The model architecture comprises three parts: the cost-volume-based module with intra evidential fusion(yellow block), the transformer-based module(blue block), and the inter evidential fusion module(green block). The framework leverages evidential estimation to accomplish two-stage fusion and generate both aleatoric and epistemic uncertainty maps along with disparity.}
\label{fig:architecture}
\end{figure*}

\section{Related Works}

\subsection{Deep Stereo Matching} 
\textbf{Cost-volume-based Deep Stereo Matching Methods.} 
Cost-volume-based deep stereo matching methods are widely used and have achieved promising results. DispNetC~\cite{sceneflow}
is the first end-to-end trainable stereo matching framework using the dot product of left and right feature maps to form a 3D cost volume. Despite of a computational-friendly way, the correlation operation cannot capture enough knowledge to obtain a satisfactory result. Following GC-Net~\cite{kendall2017end}, many works~\cite{Chang18,Zhang2019GANet,LEAStereo2020} employ 3D hourglass convolutions to aggregate 4D cost volume with the size of [height$\times$width$\times$disparity range$\times$feature] constructed by concatenating the features of stereo pairs, which demands large memory and computational complexity. Group-wise correlation is then proposed in GwcNet~\cite{gwc2019} to construct a compact cost volume and facilitates a better trade-off. ACVNet \cite{acv2022} proposes to build attention-aware cost volume to suppress redundant information and further alleviate the burden of cost aggregation.

To reduce the high computational cost and leverage more semantic and robust information, multi-scale cost volume is introduced to stereo matching. HSMNet~\cite{yang2019hierarchical} constructs a pyramid of volumes to process high-res stereo images. AANet~\cite{Xu20} is a lightweight framework with a feature pyramid and multi-scale correlation volume interaction. CFNet~\cite{Shen_2021_CVPR} constructs cascade pyramid cost volume to narrow down the disparity search range and refine the disparity map in a coarse-to-fine manner. Most recently, PCWNet~\cite{shen2022pcw} proposes a volume fusion module to directly combine multi-scale 4D volumes and calculate a multi-level loss to accelerate the convergence of the model. 

\textbf{Transformer-based Deep Stereo Matching Methods.}
With the support of attention mechanisms, transformer-based deep stereo matching methods have become another line of stereo matching research. STTR~\cite{li2021revisiting} matches pixels from a sequence-to-sequence matching perspective to impose a uniqueness constraint and avoid the construction of fixed disparity cost volume. CSTR~\cite{guo2022context} employs a plug-in module, Context Enhanced Path, to help better integrate global information and deal with hazardous regions, such as texturelessness, specularity, or transparency. 

Generally, transformer-based methods excel at modeling long-range global knowledge, but do not perform well in regions with local texture details. This encourages us to improve the overall performance by fusing transformer-based and cost-volume-based methods to capture complementary information. In this paper, we use PCWNet~\cite{shen2022pcw} and STTR~\cite{li2021revisiting} as examples for fusion, 
but other cost-volume-based and transformer-based approaches can be used as well.

\subsection{Uncertainty Estimation}
As deep learning techniques are increasingly applied in safety-sensitive real-world scenarios, the measurement of confidence thus becomes vital for distinguishing the doubtful predictions and aiding in decision making~\cite{gawlikowski2021survey}. Bayesian neural networks~\cite{kendall2017uncertainties} enable deep models with uncertainty by substituting the deterministic weight parameters with distributions. However, the computational expense of optimization is unaffordable with such a huge amount of parameters. Monte Carlo dropout (MC Dropout)~\cite{gal2016dropout} is the most well-known method to tackle the problem, which formulates dropout as Bernoulli distributed random variables to approximate the training process as variational inference. Deep ensemble methods~\cite{lakshminarayanan2017simple,chitta2018large,wen2020batchensemble} predict the overall result based on multiple models with different architectures and have gained popularity in modeling uncertainty in the past several years. Several methods are introduced to help ensemble methods become more practical, such as pruning~\cite{cavalcanti2016combining} and distillation~\cite{lindqvist2020general}. Deterministic neural network~\cite{mozejko2018inhibited,liu2020simple,van2020uncertainty} is a more efficient estimation approach, which directly computes the uncertainty of prediction distributions based on one single forward pass. For instance, evidential deep learning~\cite{sensoy2018evidential} and prior networks~\cite{malinin2018predictive} place Dirichlet priors over discrete classification predictions. Deep evidential regression~\cite{amini2020deep} proposes to extend~\cite{sensoy2018evidential} to regression tasks to estimate the parameters of the normal inverse gamma distribution over an underlying Normal distribution, which ensures explicit representation of epistemic and aleatoric uncertainties. Considering a multi modalities setting, Ma et al.~\cite{ma2021trustworthy} uses a mixture of the normal inverse gamma distribution (MoNIG) to allow a trustworthy regression which characterizes both modality-specific and integrated uncertainties.

Several works seek to incorporate uncertainty estimation into stereo tasks. UCSNet~\cite{cheng2020deep} proposes calculating adaptive thin volume with an uncertainty-aware cascaded design in a multi-view stereo setting. Inspired by UCSNet, CFNet~\cite{Shen_2021_CVPR} employs uncertainty estimation to adjust the disparity search range adaptively. In contrast to the previous methods using variance-based uncertainty, Wang et al.~\cite{wang2022uncertainty} first applies deep evidential learning to predict uncertainties of disparity map in stereo matching. In this paper, we further extend deep evidential learning to fully utilize the multi-level knowledge in stereo matching task with both intra and inter evidential fusion strategies.

%% file: 3_Methods.tex
\section{Method}
This section explains the proposed Evidential Local-global Fusion (ELF) framework based on uncertainty estimation for stereo matching. As illustrated in Figure~\ref{fig:architecture}, our network architecture  is divided into three parts: the cost-volume-based module with intra evidential fusion, the transformer-based module, and the inter evidential fusion module. Given a stereo pair, pyramid combination network with intra evidential fusion and trustworthy stereo transformer respectively predict the distribution parameters $\{\delta_{local}, \gamma_{local}, \alpha_{local}, \beta_{local}\}$ and $\{\delta_{global}, \gamma_{global}, \alpha_{global}, \beta_{global}\}$. Then, disparity, aleatoric uncertainty and epistemic uncertainty are deduced from the syncretic $\{\delta, \gamma, \alpha, \beta\}$, which are obtained by inter evidential fusion module based on the multi-view mixture of the normal-inverse gamma distribution.

\subsection{Evidential Deep Learning for Stereo Matching}

\subsubsection{Background and Uncertainty Loss}
Stereo matching strives to estimate the disparity between the given stereo pair for each pixel. From the viewpoint of evidential learning, every disparity $d$ is drawn from a normal distribution, but with unknown mean and variance $(\mu, \sigma^2)$. To model the distribution, $\mu$ and $\sigma^2$ are assumed to be drawn from normal and inverse-gamma distributions respectively:
\begin{align}
d \sim \mathcal{N} (\mu, \sigma^2),\ \mu \sim \mathcal{N} (\delta, \sigma^2\gamma^{-1}),\ \sigma^2 \sim \Gamma^{-1}(\alpha, \beta),
\end{align}
where $\Gamma(\cdot)$ denotes the gamma function, $\delta\in\mathbb{R}$, $\gamma>0$, $\alpha>1$, $\beta>0$. 

With the assumption that the mean and the variance are independent, the posterior distribution $q(\mu,\sigma^2)=p(\mu, \sigma^2 |d_1, ..., d_N)$ can be formulated as a normal-inverse gamma distribution $\text{NIG}(\delta, \gamma, \alpha, \beta)$. Amini et al.~\cite{amini2020deep} then define the total evidence $\Phi=2\gamma+\alpha$ to measure the confidence of the predictions. The disparity $d$, aleatoric uncertainty $al$ and epistemic uncertainty $ep$ can be derived as
\begin{align}
\begin{split}
d = \mathbb{E}(\mu) &= \sigma,\ al = \mathbb{E}(\sigma^2) = \frac{\beta}{\alpha-1}, \\
ep & = \text{Var}(\mu) = \frac{\beta}{\gamma(\alpha-1)},  
\end{split}
\end{align}
During training, the loss function $\mathcal{L}^{N}$ can be defined as the negative logarithm of model evidence,
\begin{align}
\begin{split}
\mathcal{L}^{N}(w) &= \frac{1}{2}\log(\frac{\pi}{\gamma})-\alpha\log(\Omega)+ \\(\alpha+\frac{1}{2})&\log((y-\delta)^2\gamma+\Omega)+\log\left(\frac{\Gamma(\alpha)}{\Gamma(\alpha+\frac{1}{2})}\right),
\end{split}
\end{align}
where $\Omega = 2\beta(1+\gamma)$, $w$ denotes the set of the estimated distribution parameters.

To reduce the evidence where prediction is incorrect, a regularization term is introduced,
\begin{align}
\mathcal{L}^{R}(w) = |d^{gt}-\mathbb{E}(\mu_i)|\cdot \Phi = |d^{gt}-\delta|\cdot (2\gamma+\alpha),
\end{align}
where $d^{gt}$ is the ground truth disparity map.

To extend deep evidential learning to the dense stereo matching task, the total uncertainty loss $\mathcal{L}^{U}$ can be defined as the expectation over all the pixels,
\begin{align}
\mathcal{L}^{U}(w) = \frac{1}{N}\sum_{0}^{N-1} \left(\mathcal{L}_{i}^{N}(w) +\tau\mathcal{L}_{i}^{R}(w)\right),
\end{align}
where $\tau>0$ controls the degree of regularization, $N$ denotes the total number of  pixels.

\begin{figure}[ht]
\centering
\includegraphics[scale=0.20]{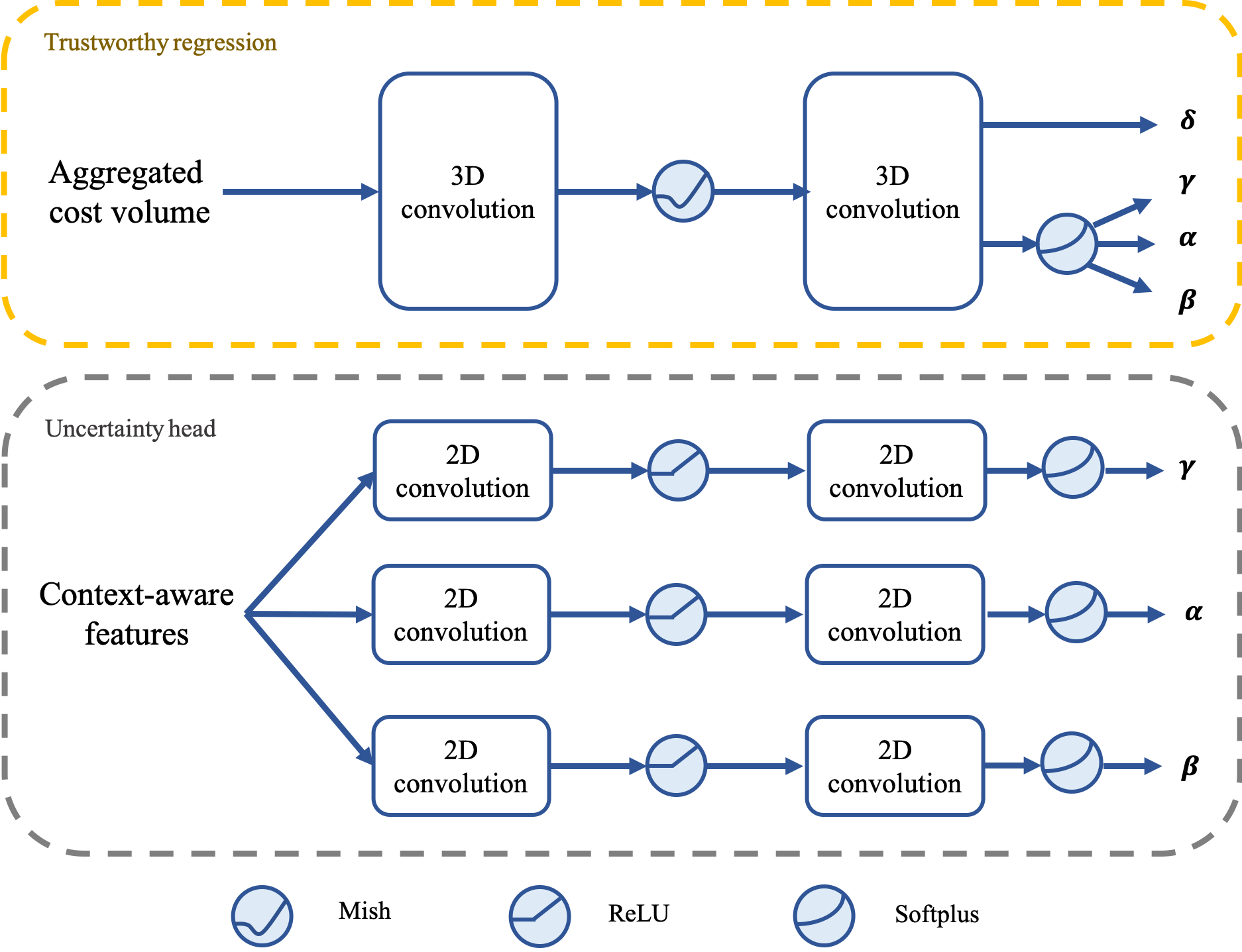}
\caption{Overview of the trustworthy regression in cost-volume-based stereo matching(top) and uncertainty head in transformer-based stereo matching(bottom). Uncertainty head only predicts $\gamma$, $\alpha$, and $\beta$, whereas $\delta$ is generated by optimal transport module.}
\label{fig:uncertainty_head}
\end{figure}

\subsubsection{Uncertainty Estimation in Stereo Matching}

\textbf{Uncertainty estimation in cost-volume-based stereo matching.} Cost-volume-based stereo matching networks have five typical main modules in detail: shared-weights feature extraction, cost volume construction, 3D cost aggregation, disparity regression, and disparity refinement. To estimate the parameters of NIG distribution rather than merely predicting the disparity map, we substitute the disparity regression module into a trustworthy regression with multi-channel output and keep the remaining modules unchanged. Shown in Figure~\ref{fig:uncertainty_head} (top), considering the computational complexity of 3D convolution operation, the proposed trustworthy regression module employs a single branch of two 3D convolutions with Mish activation~\cite{misra2019mish} and the up-sample method to output a 4-channel volume $V_{output} \in \mathbb{R}^{D_{max}\times H\times W\times 4}$, where $H$ and $W$ are the height and width of the input stereo pairs, and $D_{max}$ is the max value of disparity candidates. The distribution parameters can be computed as follows:
\begin{align}
V_{\delta}, V_{\gamma}, V_{\alpha}, V_{\beta} &= \text{Split}(V_{out}, dim=-1), \\
p &= \text{Softmax}(V_{\delta}), \\
\delta = \sum_{k=0}^{D} k &\cdot p_k,\ \text{logit}_i = \sum_{k=0}^{D} V_{i} \cdot p_k,
\end{align}
where k denotes disparity level, p denotes probability, and $i \in \{\gamma, \alpha, \beta\}$. Then a Softplus activation is applied on $\text{logits}_i$ to generate $\gamma, \alpha, \beta$. 

\textbf{Uncertainty estimation in transformer-based stereo matching.} Transformer-based stereo matching networks take advantage of cross- and self-attention mechanisms to predict the expectation of disparity $\delta$ and occlusion probability $p_{occ}$ without fixed-disparity cost volume and 3D convolutions. The Stereo Transformer outputs the attention weights $w_{attn}$, then the optimal transport module regresses the disparity and calculates occlusion probability based on the cost matrix, which is set as $-w_{attn}$. On top of this, we add an uncertainty head consisting of two 2D convolution blocks and Softplus activation(shown in Figure~\ref{fig:uncertainty_head} (bottom)) to generate the parameters $\gamma, \alpha, \beta$ from the concatenation of left and right context-aware features converted by the Stereo Transformer. To obtain better-calibrated results, we utilize the context adjustment layer in accordance with STTR~\cite{li2021revisiting} to refine the parameter $\delta$ and the occlusion probability $p_{occ}$.

\subsection{Fusion Strategy based on Evidence}

We adopt the fusion strategy with the mixture of normal-inverse gamma distribution (MoNIG)~\cite{ma2021trustworthy} for its excellent mathematical properties to perform both intra evidential fusion and inter evidential fusion. Specifically, given $M$ sets of parameters of NIG distributions, the MoNIG distribution can be computed with the following operations:
\begin{align}
\begin{split}
\text{MoNIG}(\delta, \gamma, \alpha, \beta) =& \text{NIG}(\delta_1, \gamma_1, \alpha_1, \beta_1) \oplus \\
\text{NIG}(\delta_2, \gamma_2, \alpha_2, \beta_2) \oplus &\cdots \oplus \text{NIG}(\delta_M, \gamma_M, \alpha_M, \beta_M),
\end{split}
\end{align}
where $\oplus$ represents the summation operation of two NIG distributions, which is defined as
\begin{align}
\begin{split}
\text{NIG}(\delta, \gamma, \alpha, \beta) \triangleq \text{NIG}(\delta_1, \gamma_1, \alpha_1, \beta_1) \oplus \text{NIG}(\delta_2, \gamma_2, \alpha_2, \beta_2),
\nonumber
\end{split}
\end{align}
where
\begin{align}
\begin{split}
\delta &= (\gamma_1 + \gamma_2)^{-1}(\gamma_1\delta_1 + \gamma_2\delta_2),\\
\gamma &= \gamma_1 + \gamma_2,\ \alpha = \alpha_1 + \alpha_2 +\frac{1}{2},\\
\beta &= \beta_1 + \beta_2  + \frac{1}{2}\gamma_1(\delta_1-\delta)^2 + \frac{1}{2}\gamma_2(\delta_2-\delta)^2.
\end{split}
\end{align}
The parameter $\delta$ of the combined distribution is the summation of $\delta_1$ and $\delta_2$ weighted by $\gamma$, which measures the confidence level of the expectation. The final $\beta$ is defined as not only the sum of $\beta_1$ and $\beta_2$, but also the variance between the combined distribution and each individual distribution, as it provides insight into aleatoric and epistemic uncertainties simultaneously. 

\subsubsection{Intra Evidential Fusion of Cost-volume-based Stereo Matching}

Multi-scale cost volume has been frequently used in stereo matching tasks to exploit the features from different layers of the extractor. We construct three levels of the group-wise correlation volumes with $\frac{1}{16}$, $\frac{1}{8}$, $\frac{1}{4}$ scaled features and employ the cost volume fusion module~\cite{shen2022pcw} to early combine the knowledge from multi-scale receptive field. These feature maps contain coarse-to-fine semantic information such as textures, boundaries, and regions. Then we apply three branches of 3D cost aggregation and trustworthy regression module to generate parameters of NIG distributions $(\delta_i,\gamma_i,\alpha_i,\beta_i)$, where $i\in \{1, 2, 3 \}$. Intra evidential fusion module integrates three NIG distributions into one distribution as the final pyramid combined result
\begin{align}
\begin{split}
\text{MoNIG}(\delta_{local},\gamma_{local},\alpha_{local},\beta_{local}) = & \\ \text{NIG}(\delta_1, \gamma_1, \alpha_1, \beta_1) \oplus \cdots \oplus \text{NIG}(\delta_3, &\gamma_3, \alpha_3, \beta_3).
\end{split}
\end{align}
The uncertainty-aware fusion strategy equips our framework with the ability to integrate reliable outputs from multi-scale features.  

\subsubsection{Inter Evidential Fusion between Cost-volume-based and Transformer-based Stereo Matching}

The intrinsic inductive bias of locality of convolutions makes cost-volume-based stereo matching models easy to model local features, whereas the transformer-based models capitalize on the long-range dependencies of attention mechanism to capture global information. The different foci of the two methods lead to the difference in strengths and weaknesses for predicting disparities and are likely to complement one another in some instances. The inter evidential fusion with MoNIG distribution provides an elegant and computationally efficient mechanism to merge two predictions into one. We apply the fusion strategy based on uncertainty to obtain the final distribution
\begin{align}
\begin{split}
\text{MoNIG}(\delta,&\gamma,\alpha,\beta) = \text{MoNIG}(\delta_{local},\gamma_{local},\alpha_{local},\beta_{local}) \\ 
\oplus & \text{NIG}(\delta_{global},\gamma_{global},\alpha_{global},\beta_{global}).
\end{split}
\end{align}

\subsection{Loss}

We compute the uncertainty losses on local outputs, global outputs and final combined outputs, denoted as $\mathcal{L}^{U}(w_{local})$, $\mathcal{L}^{U}(w_{global})$ and $\mathcal{L}^{U}(w)$, respectively. In the transformer-based stereo matching module, we obtain the attention weights and occlusion probability $p_{occ}$ as well. Besides the uncertainty loss, we adopt the same loss functions as STTR~\cite{li2021revisiting}, relative response loss $\mathcal{L}^{RR}(w_{attn})$ to maximize the attention on the true target location and binary-entropy loss $\mathcal{L}^{BE}(p_{occ})$ to supervise occlusion map. The overall loss function is described as
\begin{align}
\begin{split}
\mathcal{L} &=  \mathcal{L}^{U}(w_{local}) + \lambda_1\mathcal{L}^{U}(w_{global}) \\
+ &\lambda_2\mathcal{L}^{U}(w) + \lambda_3\mathcal{L}^{RR}(w_{attn}) + \lambda_4\mathcal{L}^{BE}(p_{occ}),
\end{split}
\end{align}
where $\lambda_i$, $i=1,2,3,4$ control  the loss weights.

%% file: 4_Experiments.tex
\begin{table*}[ht]
\centering
\caption{Ablation study on Scene Flow~\cite{sceneflow}.}
\scalebox{0.9}{
\begin{tabular}{cccc|cc}
\hline
\multicolumn{4}{c|}{Components} & \multicolumn{2}{c}{Scene Flow} \\ \hline
STTR~\cite{li2021revisiting} & Uncertainty & \begin{tabular}[c]{@{}c@{}}Inter Evidential \\ Fusion\end{tabular} & \begin{tabular}[c]{@{}c@{}}Intra Evidential \\ Fusion\end{tabular} & EPE(px)$\downarrow$ & D1-1px(\%)$\downarrow$   \\ \hline\hline
 $\checkmark$ &  &  &  & 0.42 & 1.37  \\
 $\checkmark$ & $\checkmark$  &  &  & 0.41 & 1.31  \\
 $\checkmark$ & $\checkmark$  &$\checkmark$  &  & 0.38 & 1.32 \\
 $\checkmark$ & $\checkmark$  &$\checkmark$  &$\checkmark$  & \textbf{0.33} & \textbf{1.28}  \\ \hline
\end{tabular}}
\label{table: ablation_study}
\end{table*}

\section{Experiments}
In this section, we evaluate the proposed ELFNet on various datasets including Scene Flow~\cite{sceneflow}, KITTI 2012 \& KITTI 2015~\cite{kitti2012,kitti2015} and Middlebury 2014~\cite{scharstein2014high}. Additionally, we conduct uncertainty analyses to explore the relationship between model performance and uncertainties.

\subsection{Datasets and Evaluation Metrics}
$\textbf{Scene Flow FlyingThings3D subset}$ \cite{sceneflow} is a large-scale synthetic dataset of random objects. The subset provides about 25,000 stereo frames with the resolution of 960$\times$540 and corresponding sub-pixel ground truth disparity maps and occlusion regions.

$\textbf{KITTI 2012 \& KITTI 2015}$ \cite{kitti2012,kitti2015} are collected from the real-life driving scenario. KITTI 2012 provides 194 training and 195 testing images pairs, and KITTI 2015 provides 200 training and 200 testing image pairs. The resolution of KITTI 2012 and KITTI 2015 stereo pairs are 1226$\times$370 and 1242$\times$375. Both datasets provide sparse disparity maps.

$\textbf{Middlebury 2014}$ \cite{scharstein2014high} is an indoor dataset with 15 training image pairs and 15 testing image pairs. It contains high resolution stereo pairs in three different resolutions, and we select the quarter-resolution ones.

$\textbf{Evaluation metrics.}$ We use end point error (EPE), percentage of disparity outliers by 1px or 3px (D1-1px / D1-3px) and percentage of errors larger than 3px (3px Err) as evaluation metrics for disparity prediction.

\subsection{Implementation Details}
Our ELF framework is inherently compatible with all transformer-based and multi-scale cost-volume-based models. For our experiments, we set STTR~\cite{li2021revisiting} as the transformer-based part and PCWNet~\cite{shen2022pcw} as the cost-volume-based part for its good balance between accuracy and generalization. We train the model in an end-to-end manner using AdamW optimizer with a weight decay of 1e-4. We pre-train on Scene Flow FlyingThings3D subset~\cite{sceneflow} for 16 epochs with the initial learning rate of 2e-4 for the transformer-based part and 2e-3 for the cost-volume-based part.
After epochs 4, 8, 10, and 12, we decrease the learning rate by a factor of 2. 
We use 6 self- and cross-attention layers with 4 heads for the transformer-based part and set the maximum disparity $D = 192$ for the cost-volume-based part. The regularization weight in the uncertainty loss $\tau = 0.5$. The weights of four outputs are set as  $\lambda_1=1.0$, $\lambda_2=2.0$, $\lambda_3=1.0$, $\lambda_4=1.0$ empirically. We apply data augmentations to stimulate real-world scenarios during training, including Gaussian noise, brightness/contrast shift, and random size cropping. 
All the experiments are conducted on a NVIDIA RTX 3090 GPU.

\subsection{Comparison with State-of-the-art}

\begin{table}[ht]
\centering
\caption{Comparison with state-of-the-art on Scene Flow~\cite{sceneflow}.}
\scalebox{0.78}{\begin{tabular}{l|cc|cc}
\hline
 & \multicolumn{2}{c|}{Disparity \textless 192} & \multicolumn{2}{c}{All Pixels} \\ \cline{2-5} 
 & EPE(px)$\downarrow$ & D1-1px(\%)$\downarrow$ & EPE(px)$\downarrow$ & D1-1px(\%)$\downarrow$ \\ \hline\hline
PSMNet~\cite{Chang18} &  0.95 & 2.71 &  1.25 & 3.25  \\
GwcNet~\cite{gwc2019} &  0.76 & 3.75  &  3.44 & 4.65  \\
CFNet~\cite{Shen_2021_CVPR} &  0.70 & 3.69  &  1.18 & 4.26 \\
PCWNet~\cite{shen2022pcw} &  0.85 & 1.94  &  0.97 & 2.48 \\
GANet~\cite{Zhang2019GANet} &  0.48 &  4.02  &  0.97 & 4.89  \\
STTR~\cite{li2021revisiting} &  0.42 & 1.37  &  0.45 & \textbf{1.38}  \\
CSTR~\cite{guo2022context} & 0.41 & 1.41  &  0.45 & 1.39  \\ \hline
ELFNet(Ours)  & \textbf{0.33} & \textbf{1.28}  &  \textbf{0.40} & 1.39 \\ \hline
\end{tabular}}
\label{table: pre-train}
\end{table}

To evaluate the effectiveness of the proposed approach, we compare it to several state-of-the-art methods, including PSMNet~\cite{Chang18}, GwcNet~\cite{gwc2019}, CFNet~\cite{Shen_2021_CVPR}, PCWNet~\cite{shen2022pcw}, GANet~\cite{Zhang2019GANet}, STTR~\cite{li2021revisiting} and CSTR~\cite{guo2022context}. Table~\ref{table: pre-train} presents a comparison of our method with the previous state-of-the-art methods on Scene Flow~\cite{sceneflow}. Our proposed method outperforms all other methods in both EPE and D1-1px metrics with different settings. Notably, our approach demonstrates an improvement of 19.5\% (0.33 v.s. 0.41) in EPE and 9.2\% (1.28 v.s. 1.41) in D1-1px compared to the best performing method,  CSTR~\cite{guo2022context}, in the Disparity $<$ 192 setting.

In the All Pixels setting, our approach demonstrates a reduced EPE over the current state-of-the-art method by 11.2\% from 0.45 to 0.40. Overall, our ELFNet outperforms the cost-volume-based and transformer-based models in disparity estimation accuracy by leveraging their advantages in a trustworthy manner. Meanwhile, it also maintains occlusion estimation from transformer. It achieves comparable occlusion estimation with occlusion intersection over union score 0.98  compared with 0.97 by STTR~\cite{li2021revisiting}.

\begin{table*}[ht]
\centering
\caption{Cross-domain evaluation without \textit{fine-tuning} on Middleburry 2014~\cite{scharstein2014high}, KITTI 2012~\cite{kitti2012} and KITTI 2015~\cite{kitti2015}.}
\scalebox{0.8}{
\begin{tabular}{l|cc|cc|cc}
\hline
  & \multicolumn{2}{c|}{Middlebury 2014} & \multicolumn{2}{c|}{KITTI 2012} & \multicolumn{2}{c}{KITTI 2015}  \\ \cline{2-7} 
& EPE(px)$\downarrow$ & 3px Err(\%)$\downarrow$ & EPE(px)$\downarrow$ & D1-3px(\%)$\downarrow$ & EPE(px)$\downarrow$ & D1-3px(\%)$\downarrow$  \\ \hline\hline
PSMNet~\cite{Chang18}   & 3.05 & 13.0 & 3.46 & 25.9 & 6.59 & 16.3 \\
GwcNet~\cite{gwc2019}   & 1.89 & 8.95 & 1.68 & 11.7 & 2.21 & 12.2 \\
CFNet~\cite{Shen_2021_CVPR} & \textbf{1.69} & 7.73 & \textbf{0.96} & \textbf{4.74} & 2.27 & \textbf{5.76} \\
PCWNet~\cite{shen2022pcw}   & 2.17 & 9.09 & 1.32 & 5.37 & 1.88 & 6.03 \\
STTR~\cite{li2021revisiting}  & 2.33 & \underline{6.19} & 1.82 & 6.96 & \textbf{1.50} & 6.40 \\ \hline
ELFNet (Ours)   & \underline{1.79} & \textbf{5.72} & \underline{1.18} & \textbf{4.74} & \underline{1.57}  & \underline{5.82} \\ \hline
\end{tabular}}
\label{table: cross_domain}
\end{table*}

\begin{figure*}[htb] 
\centering
\subfigure[Aleatoric uncertainty curve]{
\centering
\begin{minipage}[b]{0.25\textwidth}
\includegraphics[width=1\textwidth]{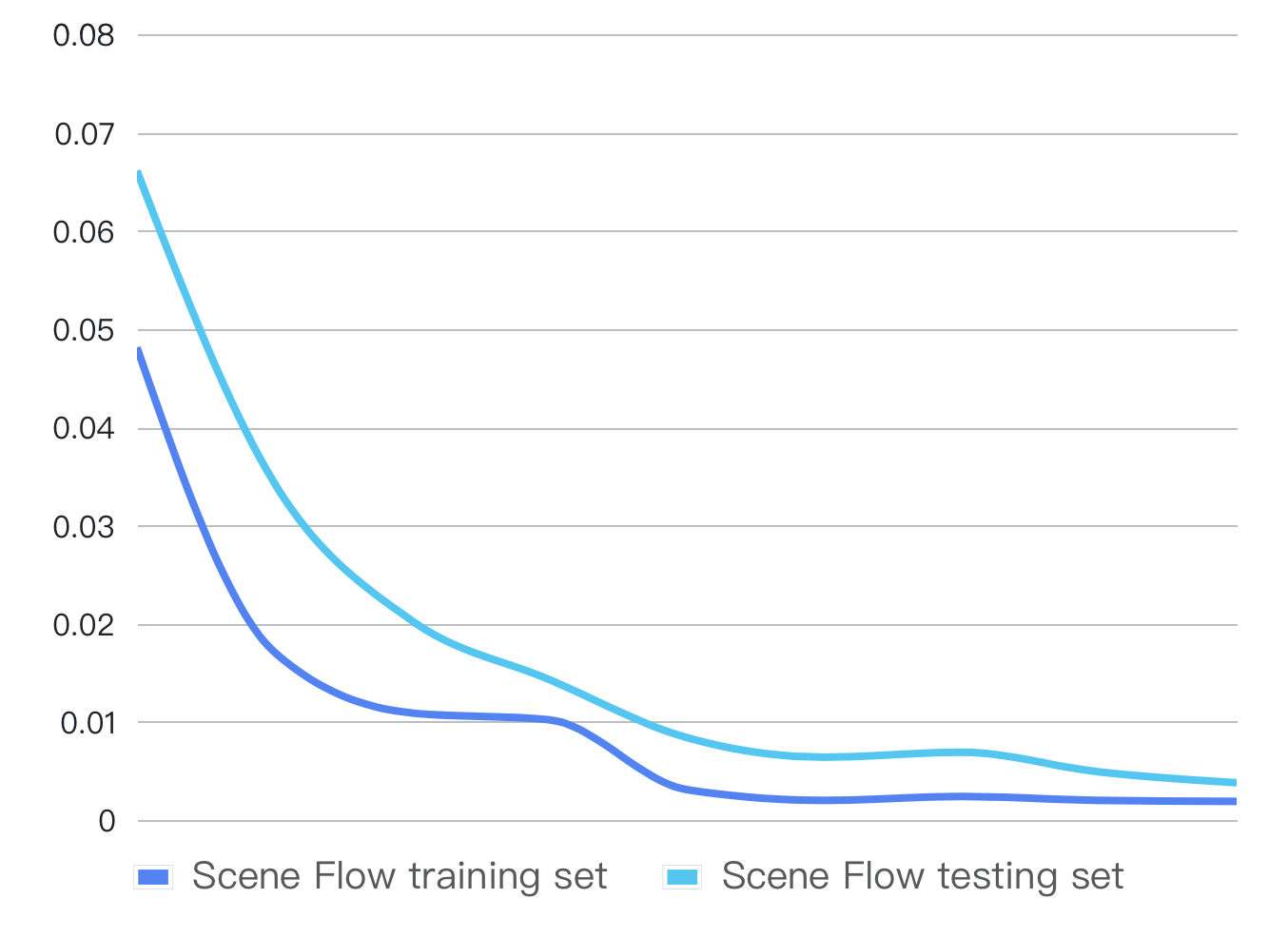}
\label{al_curve}
\end{minipage}
}
\centering
\subfigure[Epistemic uncertainty curve]{
\centering
\begin{minipage}[b]{0.25\textwidth}
\includegraphics[width=1\textwidth]{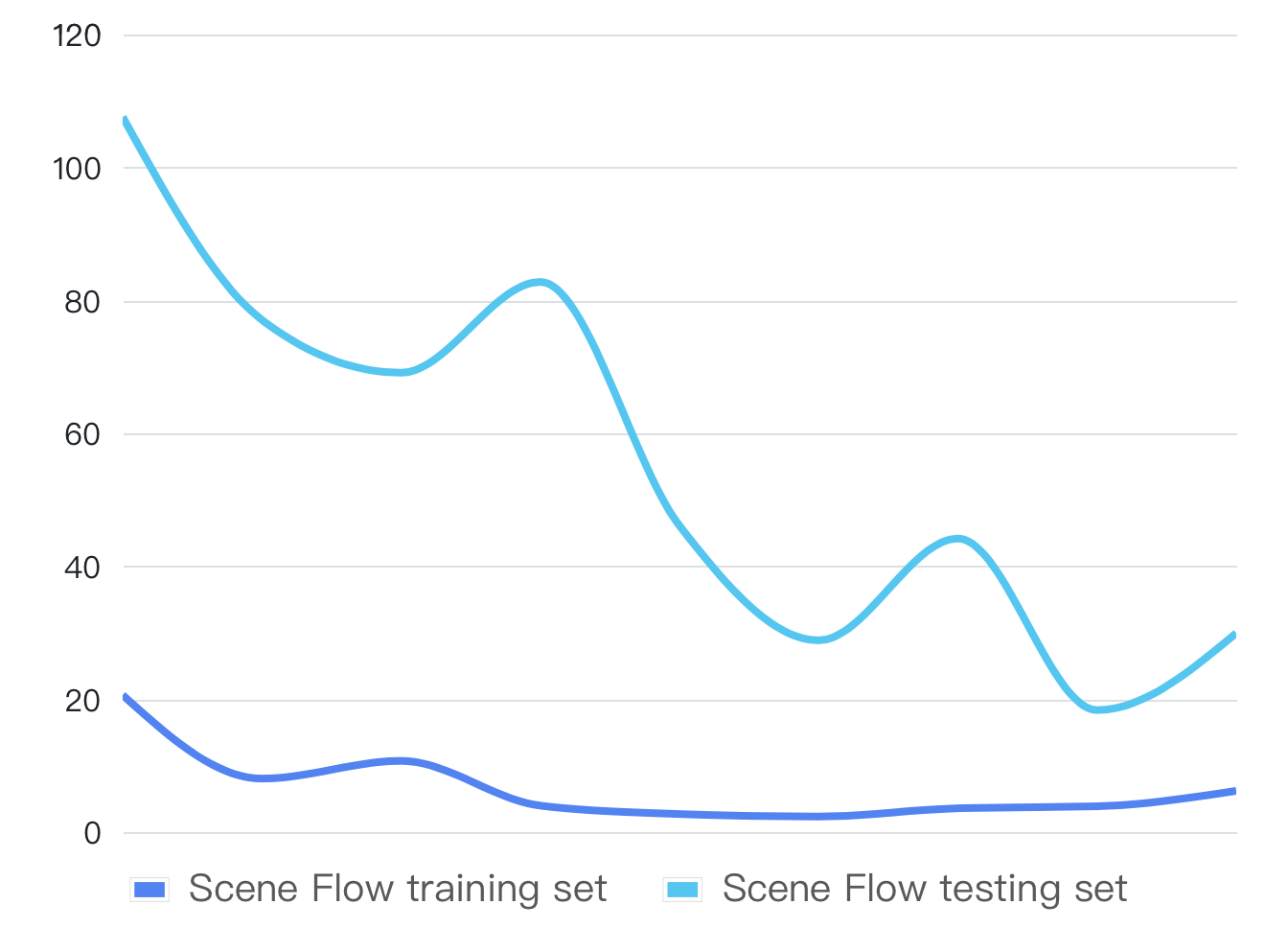}
\label{ep_curve}
\end{minipage}
}
\centering
\subfigure[Scene Flow~\cite{sceneflow}]{
\begin{minipage}[b]{0.22\textwidth}
\centering
\includegraphics[width=0.9\textwidth]{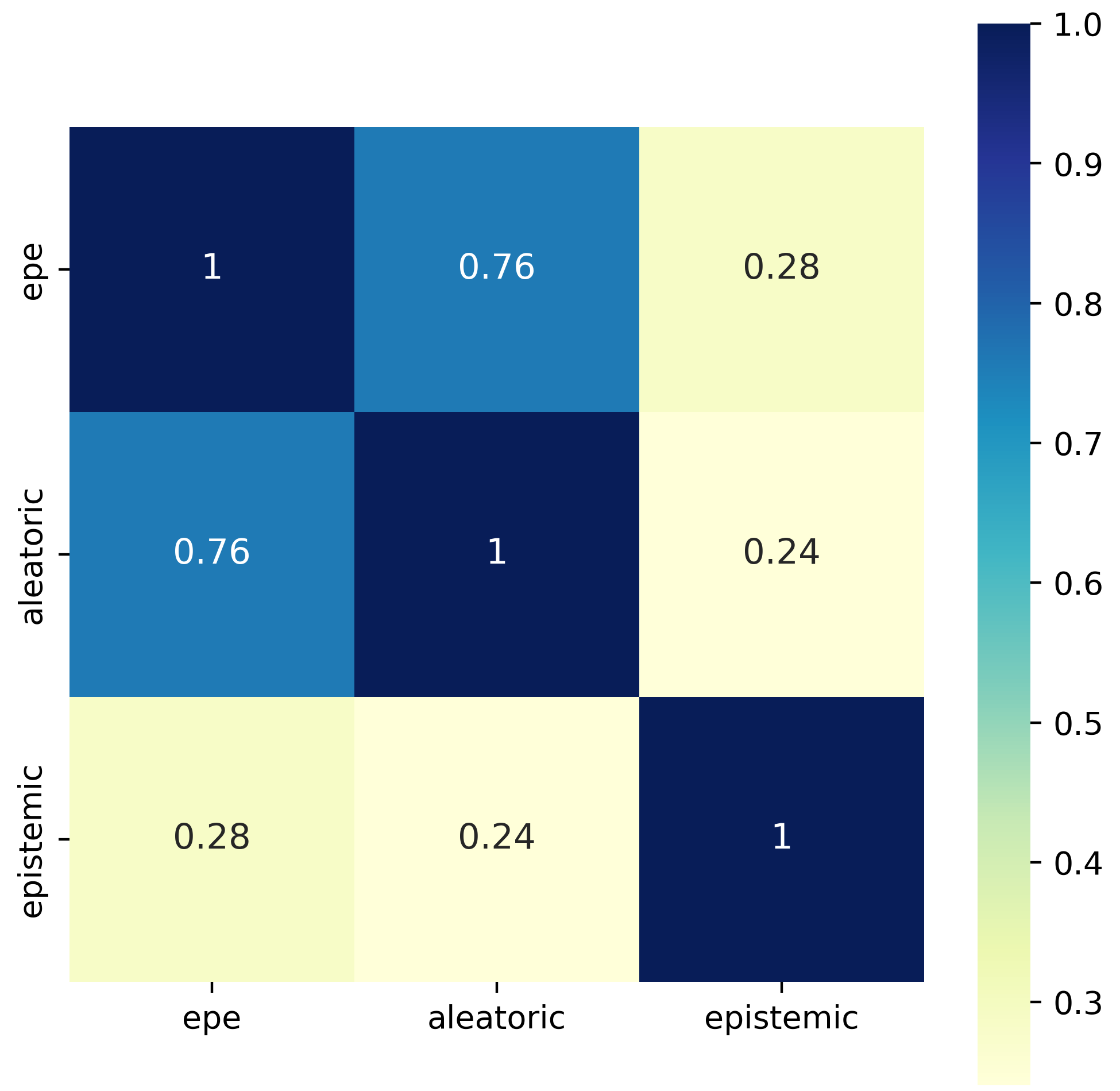}
\label{sceneflow_heatmap}
\end{minipage}
}
\centering
\subfigure[KITTI 2015~\cite{kitti2015}]{
\centering
\begin{minipage}[b]{0.22\textwidth}
\includegraphics[width=0.9\textwidth]{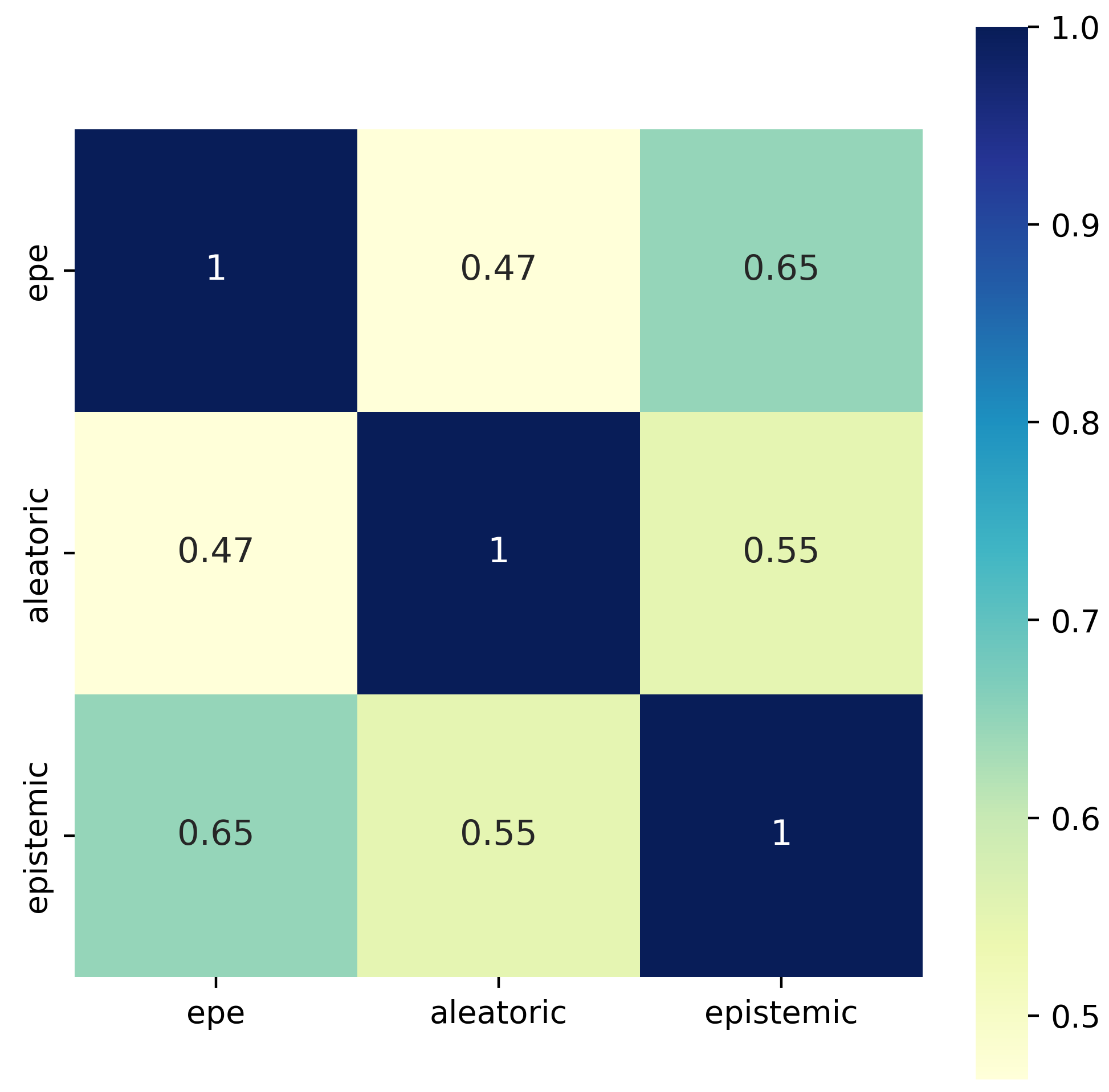}
\label{kitti_heatmap}
\end{minipage}
}
\caption{Uncertainty analysis. (a) and (b) are the training curve of two uncertainties on the Scene Flow~\cite{sceneflow}; (c) and (d) are the heat maps of the Pearson correlation analysis between the EPE and the uncertainties.}
\label{figure: heatmap}
\end{figure*}

\subsection{Ablation Studies}
To verify the effects of modules in our framework, we provide quantitative results in Table~\ref{table: ablation_study}. The proposed ELF framework has three main designs, including uncertainty estimation, inter evidential fusion, and intra evidential fusion. Uncertainty estimation enables the model to predict the epistemic and aleatoric uncertainties respectively; the inter evidential fusion combines the information of the transformer-based and the cost-volume based models; and the intra evidential fusion leverages different levels of local knowledge inside the cost-volume-based model. 

The ablation study indicates that all three designs are indispensable, and the evidential fusion parts play a crucial role in boosting performance. To be specific, on Scene Flow~\cite{sceneflow}, we observe that STTR~\cite{li2021revisiting} with uncertainty estimation module shows a comparable EPE score (0.42 v.s. 0.41) compared with the baseline. When employing the inter evidential fusion module, the model achieves a 0.03 improvement in terms of the EPE. With the additional intra evidential fusion module of the cost-volume-based part, the EPE of the model boosts by a remarkable 0.05. Overall, the EPE is reduced by 21.4\% (0.42 v.s. 0.33) with our ELF framework on Scene Flow~\cite{sceneflow}. ELFNet also outperforms the baseline by 6.6\% (1.28 v.s. 1.37) for the D1-1px metric.

\subsection{Fusion Strategies Comparison}
\begin{figure*}[t]
\centering
\includegraphics[scale=0.310]{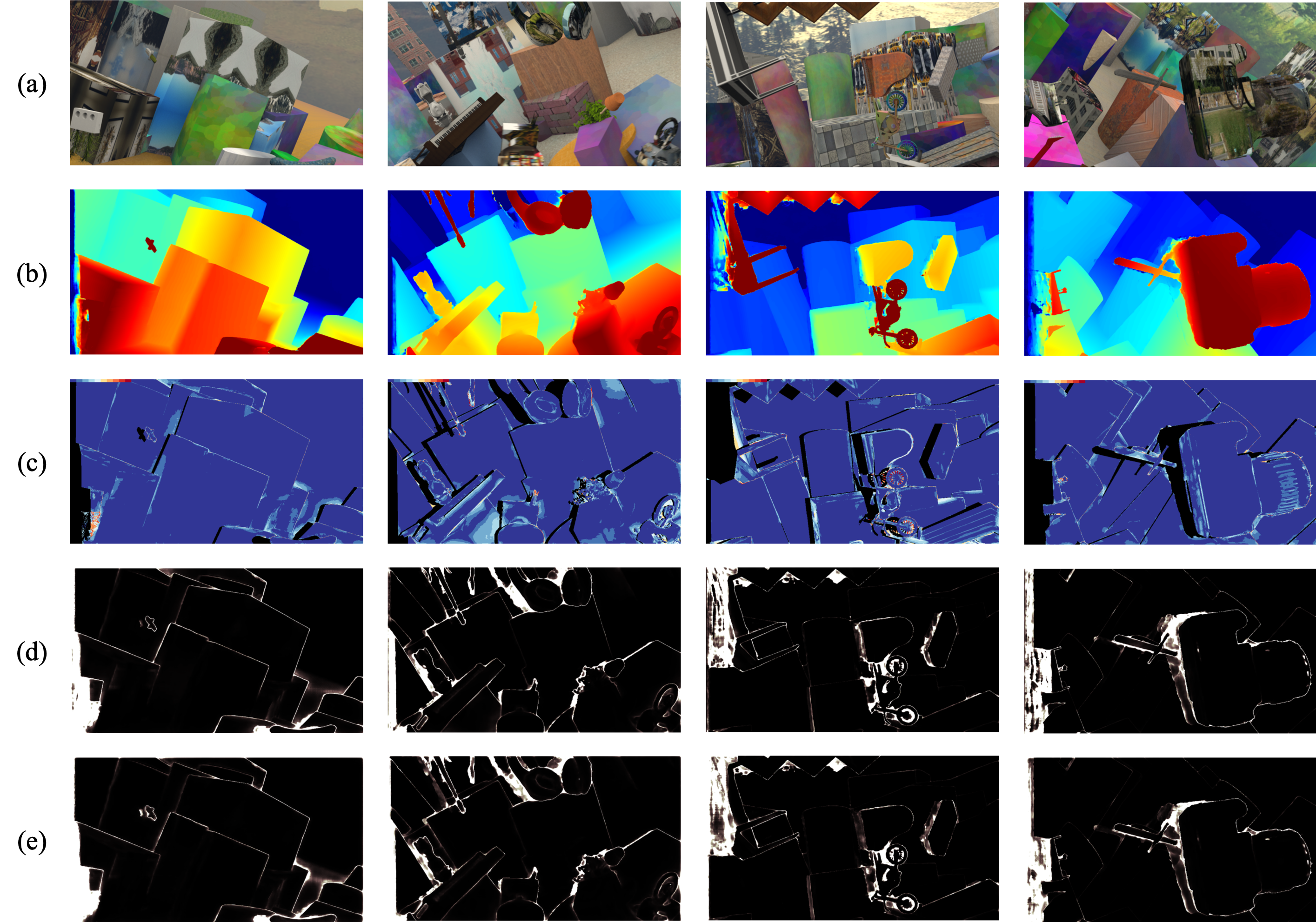}
\caption{Visualization on Scene Flow of the proposed ELFNet. (a)  The left image. (b) The estimated disparity map. (c) The error map. (d) The epistemic uncertainty map. (e)  The aleatoric uncertainty map.}
\label{fig:sceneflow_test}
\end{figure*}

\begin{table}[ht]
\centering
\caption{Fusion strategy comparison on Scene Flow~\cite{sceneflow}. Avg.: take the average of the separate estimation as the final disparity map. Conv.: add a 2D convolution layer as the late fusion module, and finetune for 5 epochs with the former parts frozen.}
\scalebox{0.85}{
\begin{tabular}{l|cc}
\hline
 &EPE(px)$\downarrow$ & D1-1px(\%)$\downarrow$ \\ \hline\hline
PCWNet~\cite{shen2022pcw} & 0.85 & 1.94 \\
STTR~\cite{li2021revisiting}  & 0.42 & 1.37 \\ \hline
Avg.  & 0.65 & 1.75 \\
Conv. & 0.42 & 1.38  \\
ELFNet(Ours) & \textbf{0.33} & \textbf{1.28}  \\ \hline
\end{tabular}}
\label{table: fusion comparison}
\end{table}

The proposed ELF framework can be viewed as a powerful late fusion strategy. To further verify the fusion performance, we compare other late fusion strategies with our ELFNet. In Table~\ref{table: fusion comparison}, we observe that simply computing the average of the output disparity maps or employing the convolution layer as a late fusion cannot achieve satisfactory results and may perform worse than STTR~\cite{li2021revisiting}. In contrast, ELFNet combines the cost-volume-based model and transformer-based model effectively  with improved results.

\subsection{Cross-domain Generalization}
We conduct experiments to prove that our pre-trained model on the synthetic Scene Flow dataset~\cite{sceneflow} can achieve strong cross-domain generalization in the zero-shot setting. As shown in Table~\ref{table: cross_domain}, our proposed ELFNet presents comparable generalization ability with the existing state-of-art models on the real-world datasets. Specially, compared with PCWNet~\cite{shen2022pcw} baseline, ELFNet brings a 17.5\% EPE score (1.79 v.s. 2.17) and a 37.1\% 3px Err score (5.72 v.s. 9.09) improvement on Middlebury 2014~\cite{scharstein2014high}, and a 10.6\% EPE score (1.18 v.s. 1.32) and 11.7\% D1-3px score (4.74 v.s. 5.37) improvement on KITTI 2012~\cite{kitti2012}. On the KITTI 2015~\cite{kitti2015}, our ELFNet also achieves overall competitive generalization results. ELFNet improves 9.1\% on the D1-3px metric (5.82 v.s. 6.40) compared with STTR~\cite{li2021revisiting}.

\subsection{Uncertainty Analysis}

Uncertainty estimation with deep evidential learning provides both aleatoric and epistemic uncertainties~\cite{kendall2017uncertainties} of the predictions. The aleatoric uncertainty reflects the extent to which disparity value differs from ground truth, which is related to data noise and cannot be reduced by optimization, while epistemic uncertainty represents the degree of dispersion of disparity values, which is related to model capacity. 

Figure~\ref{figure: heatmap} shows several results of uncertainty analysis. As presented in Figure~\ref{al_curve} and \ref{ep_curve}, both aleatoric and epistemic uncertainties present a general decline during training, which shows that the model assigns lower uncertainties as learning more from the data. Based on the ELFNet pretrained on the Scene Flow~\cite{sceneflow} dataset, we conduct the Pearson correlation analysis between the EPE metric and two uncertainties tested on the Scene Flow~\cite{sceneflow} and KITTI 2015~\cite{kitti2015} separately to shed light on the uncertainty estimation mechanism. The correlation heat maps are shown in Figure~\ref{sceneflow_heatmap} and \ref{kitti_heatmap}. We have two main observations. Firstly, the correlation value is consistently positive, 
which indicates that uncertainties can serve as an indicator of accuracy during inference.
Secondly, the correlation value between EPE and aleatoric uncertainty on the Scene Flow~\cite{sceneflow} dataset is higher than the corresponding value (0.76 v.s. 0.47) on the KITTI 2015~\cite{kitti2015} dataset, while the correlation value between EPE and epistemic uncertainty is lower (0.28 v.s. 0.65). It can be inferred that when dealing with out-of-domain samples, the disparity error is more closely related to epistemic uncertainty. In contrast, if the model has been trained on similar data distributions, the error is more likely to be influenced by aleatoric uncertainty.
It is worth noting that the behaviour of estimated uncertainties is still influenced by various other factors, including the model architecture, the training strategy, the intensity of data noise, and more. 

We provide the qualitative results on Scene Flow~\cite{sceneflow} in Figure~\ref{fig:sceneflow_test}. Although there is no ground truth for uncertainties, we observe that high uncertainties are assigned in the occluded and boundary regions. 
Compared with the error maps, the uncertainty maps are also active in the areas where the error occurs, such as the wheel hub of a bicycle, which suggests that uncertainty maps provide clues for estimation offset.

\subsection{Limitations}
Although ELFNet enables uncertainty estimation and brings notable improvement, the inference speed poses a main limitation. Our framework requires more time since it involves two individual parts. We will consider applying efficient methods in the further work, such as constructing adaptive and sparse cost-volumes ~\cite{lu2018sparse, sun2018pwc, duggal2019deeppruner, wang2021patchmatchnet}.

%% file: 5_Conclusion.tex
\section{Conclusion}
In this paper, we have proposed an Evidential Local-global Fusion (ELF) framework to fuse multi-view information for stereo matching reliably. We leverage deep evidential learning to estimate multi-level aleatoric and epistemic uncertainties alongside the disparity maps, which further allows a trustworthy fusion strategy based on evidence to exploit complementary knowledge. Experimental results show that our model performs well on both accuracy and generalization across different datasets.